\documentclass[letterpaper, 10 pt, conference]{ieeeconf}  

\IEEEoverridecommandlockouts                              

\usepackage{amsmath,amsfonts,bm}

\def\eqref#1{equation~\ref{#1}}

\def\1{\bm{1}}

\DeclareMathAlphabet{\mathsfit}{\encodingdefault}{\sfdefault}{m}{sl}
\SetMathAlphabet{\mathsfit}{bold}{\encodingdefault}{\sfdefault}{bx}{n}

\usepackage[english]{babel}

\usepackage{caption}
\usepackage{subcaption}
\usepackage{graphicx}
\usepackage{booktabs}
\usepackage{stfloats}
\usepackage{multirow}

\title{\LARGE \bf
Learning Spatial and Temporal Hierarchies: \\Hierarchical Active Inference for navigation \\in Multi-Room Maze Environments
}

\author{Daria de Tinguy\textsuperscript{1}, Toon Van de Maele\textsuperscript{1}, Tim Verbelen\textsuperscript{2} and Bart Dhoedt\textsuperscript{1}
\thanks{\textsuperscript{1}Department of Information Technology, University of Ghent, Ghent, Belgium,\textsuperscript{2}Verses AI, Vancouver, Canada
        {\tt\small \{Correspondence to: Daria de Tinguy <Daria.detinguy at
ugent.be>}}%
}

\begin{document}

\maketitle
\thispagestyle{empty}
\pagestyle{empty}


\begin{abstract}

Cognitive maps play a crucial role in facilitating flexible behaviour by representing spatial and conceptual relationships within an environment. The ability to learn and infer the underlying structure of the environment is crucial for effective exploration and navigation. This paper introduces a hierarchical active inference model addressing the challenge of inferring structure in the world from pixel-based observations. We propose a three-layer hierarchical model consisting of a cognitive map, an allocentric, and an egocentric world model, combining curiosity-driven exploration with goal-oriented behaviour at the different levels of reasoning from context to place to motion. This allows for efficient exploration and goal-directed search in room-structured mini-grid environments.

\end{abstract}

\section{Introduction}

The development of autonomous systems able to navigate in their environment is a crucial step towards building intelligent agents that can interact with the real world.
Developing navigation skills in artificial agents to mirror the natural navigational abilities observed in animals, enabling these agents to adeptly move autonomously through their surroundings, has been a topic of great interest in the field of robotics and artificial intelligence. Understanding complex and potentially aliased environments and effectively navigating them require both spatial hierarchy, i.e. capturing spatial structures and relationships \cite{Self-labelling}, and temporal hierarchy \cite{zakharov2020episodic} as they are essential for devising long-term navigation strategies. This has led to the exploration of various approaches, including cognitive mapping inspired by animal navigation strategies.

In the animal kingdom, cognitive mapping plays a crucial role in navigation. Cognitive maps allow animals to understand the spatial layout of their surroundings \cite{humans-cognitive-map,humans-mapping, map-graph-cognition}, remember key locations, solve ambiguities thanks to context \cite{humans-hierarchic-plan-subway} and plan efficient routes \cite{humans-hierarchic-plan-subway, humans-hierarchic-plan-clusters}. By leveraging cognitive mapping strategies, animals can successfully navigate complex environments, adapt to changes, and return to previously visited places. 

Several approaches have been proposed to learn the structure of the world in the context of navigation. \cite{CSCG} proposes a clone structured graph representation of the environment to disambiguate aliased observations. \cite{weird_HAIF} presents a deep hierarchical model based on active inference and casts structure learning as a Bayesian model reduction problem. \cite{pezzulo_Hgenerative_model} introduces a hierarchical generative model learning and recognising maze structures based on specific localisation in a global prefixed frame. 
While all these generative models aim to capture the underlying structure and dynamics of the world, these are typically limited to small simulations with discrete state and observation spaces. 

Addressing this aspect, recent approaches like G-SLAM \cite{g-slam} and Dreamer \cite{dreamer} use deep neural networks to learn generative world models from high-dimensional observations such as pixels. However, as these capture the world in a flat latent state space, these models struggle with long-term planning, especially in aliased environments.

\begin{figure}[t!]
\vskip 0.05in
\begin{center}
\centerline{\includegraphics[width=5cm]{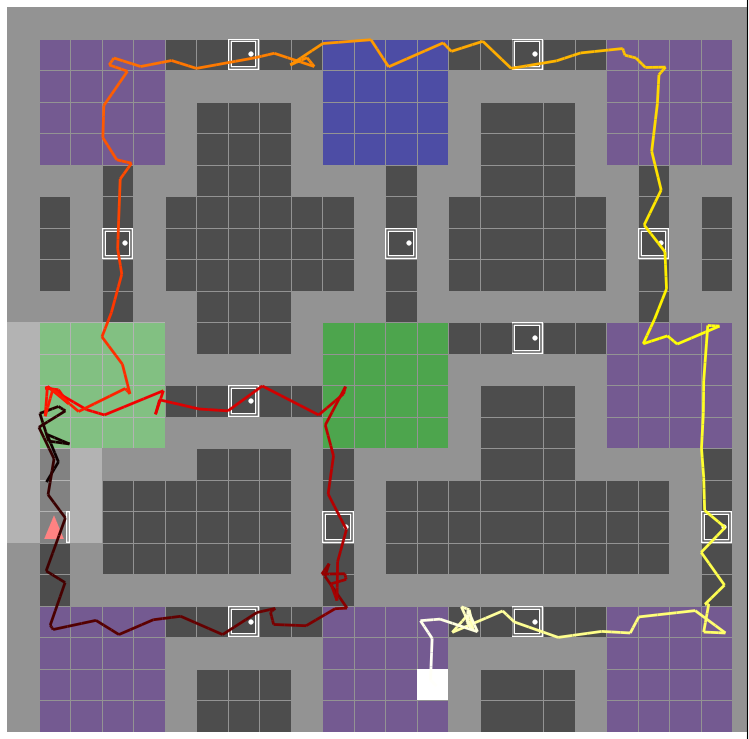}}
\caption{Example of a 3 $\times$ 3 rooms mini-grid environment and our model navigation in it during an exploration and goal-reaching task, where the starting position is the red triangle. Noise on the visualised path was added in post-processing for observing superposed visits on a single tile.}
\label{img:3x3_env}
\end{center}
\vspace{-8mm}
\end{figure}

In this paper, in order to enhance the agent's ability to navigate autonomously and intelligently we propose a pixel-based hierarchical active inference model exhibiting both spatial and temporal hierarchies. The model is geared towards learning the structure of maze mini-grid environments \cite{gym_minigrid}.
A maze consists of interconnected, visually similar rooms with variations in shape, size, and colour as depicted in Fig \ref{img:3x3_env}. Within the maze, there is a single white goal tile.
Our model consists of amortised inference models at the lower levels, which are trained on pixel data, for representing movement and pose in egocentric and allocentric reference frames respectively, combined with a graph-structured model at the top to capture the maze structure. 
The model navigation is based on a principled active inference approach, which balances goal-directed behaviour and epistemic foraging through information gain~\cite{AIF_learning}. Moreover, the planning happens at different temporal time scales at each level in the hierarchy, allowing for long-term decision-making.

Our contributions can be summarised as follows:
    \begin{itemize}
        \item We introduce a system leveraging hierarchical active inference and world modelling for achieving autonomous navigation without the necessity for task-specific training. This approach allows agents to navigate intelligently in familiar environments.
        \item Our system is built upon visual observations, which holds promise for real-world applications. 
        \item The proposed system not only learns the spatial structure of the environment but also adapts to its dynamic constraints enhancing autonomous navigation.
        \item  It creates an internal map of the entire environment, exhibiting scalability by not demanding increased computational resources with larger environments.
        \item We conduct comprehensive evaluations in a mini-grid room maze environment \cite{gym_minigrid}. Our approach demonstrates its efficiency in tasks related to exploration and goal attainment, outperforming other Reinforcement Learning (RL) models.
        \item Through quantitative and qualitative analyses, we demonstrate the effectiveness of our hierarchical active inference world model in accomplishing various tasks. Our approach exhibits resilience to aliasing and showcases its ability to learn the underlying structure of the environment.
    \end{itemize}

\begin{figure*}[t!]
\begin{center}
  \includegraphics[width=14cm]{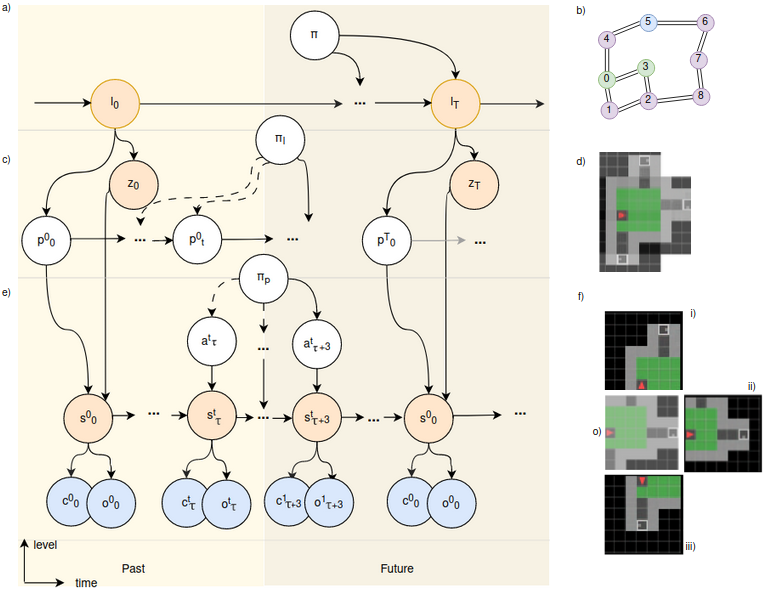}
  \caption{Our generative model unrolled in time and levels as defined in Eq.\ref{eq:HAIF}. The left shows the graphical model of the 3-layer hierarchical Active inference model consisting of a) the cognitive map, b) the allocentric model, and c) the egocentric model, each operating at a different time scale. The orange circles represent latent states that have to be inferred, the blue circles denote observable outcomes and the white circles are random variables to be inferred by planning. The right part visualises the representation at each layer. The cognitive map is represented as d) a topological graph composed of all the locations ($l$) and their connections, in which each location is stored in a distinct node. The allocentric model e) infers place representations ($z$) by integrating sequences of states ($s$) and poses ($p$), from which the room structure can be generated. The egocentric model f) imagines future observations given the current position, state ($s$), and possible actions ($a$) from this position. Here o) depicts an actual observation ($o$) and the predicted observations of the possible actions left i), forward ii), and right iii).}
  \label{img:HAIF_more}
\end{center}
\vspace{-4mm}
\end{figure*}

\section{Method}

\begin{table}[htb!]
\centering
\begin{tabular}{|c|l|}
\hline
Symbols & \multicolumn{1}{c|}{Associated Terms} \\ \hline
l & location, experience \\ \hline
z & place, room, allocentric state \\ \hline
p & pose, position \\ \hline
s & egocentric state \\ \hline
a & action \\ \hline
o & observation \\ \hline
c & collision \\ \hline
$\pi_x$ & policy, sequence of x  \\ \hline
\end{tabular}
\caption{Description of the variables used in our model}
\end{table}

The active inference framework~\cite{parr_active_2022} is built on the premise that intelligent agents minimise their surprise. An active inference agent entails an internal generative model aiming to best explain the causes of external observation and the consequences of its actions through the minimisation of surprise or prediction error, which is also known as free energy (FE). Agents minimise this quantity with respect to model parameters in learning and with respect to action in planning~\cite{AIF_learning, nav_aif}.

We propose a hierarchical generative model consisting of three layers functioning at nested timescales (see Fig~\ref{img:HAIF_more}). From top to bottom: the cognitive map, creating a coherent topological map, the allocentric model, representing space, and the egocentric model, managing motions. The system infers the environment's structure, over time, by agglomerating visual observations, creating representations of distinct places such as rooms, and progressively revealing the maze's connectivity as a graph.

\textbf{The cognitive map}: The top layer in the generative model, illustrated in Fig~\ref{img:HAIF_more}a)  functions at the coarsest time scale ($T$), each tick at this time scale corresponds to a distinct location ($l_T$) integrating the initial positions ($p^T_0$) of the place ($z^T$). These locations are depicted as nodes in a topological graph, as shown in Fig~\ref{img:HAIF_more}d). Edges between nodes are added as the agent moves from one location to another, effectively learning the maze structure. In order to maintain the spatial structure between locations, the agent keeps track of its relative rotation and translation using a continuous attractor network (CAN) as in~\cite{ratslam_hippo}. Hence the cognitive map forms a comprehensive representation of the environment, enabling the agent to navigate by formulating believes over its surroundings. 

\textbf{The allocentric model}: The middle layer, illustrated in Fig~\ref{img:HAIF_more}b), plays a crucial role in constructing a coherent understanding of the environment, denoted as $z_T$. This model functions at a finer time scale ($t$), forming a belief over the place by integrating a sequence of observations ($s^T_t$) and poses ($p^T_t$) to generate this representation \cite{GQN_origin, GQN_Toon}. Fig \ref{img:HAIF_more}e) and Fig \ref{img:room_generation} showcase the resulting place defining the environment given accumulated observations. As the agent moves from one place to another, once the current observations do not align with the previously formed prediction about the place, the allocentric model resets its place description and gathers new evidence to construct a representation of the newly discovered room ($z_{T+1}$), advancing by one tick on the coarser time scale and resetting the mid-level time scale $t$ to 0.

\textbf{The egocentric model}: The lowest layer, illustrated in Fig~\ref{img:HAIF_more}c), has the finest time scale ($\tau$). To evolve in time this model requires the prior state ($s^t_{\tau}$) and current action ($a^t_{\tau+1}$) to infer the current observation $(o^t_{\tau+1}$) \cite{generative_models_oz}. Based on its current position, the model generates possible future trajectories while considering the constraints imposed by the environment, such as the inability to pass through walls (achieved by discerning the cause-effect relationship between actions and observations). Fig \ref{img:HAIF_more}f) illustrates the current observation in the middle o) and shows the imagined potential observations if the agent were to turn left i), right iii), or move forward ii).

\textbf{Planning}:  The model operates within a hierarchical active inference scheme, planning at different time scales.
The cognitive map plays a vital role in long-term navigation by handling place connectivity, allowing the model to plan the locations to visit ($\pi$) at a high level, the poses to visit within each place ($\pi_l$) at a mid-level, and determining the best action policy ($\pi_p$) at the low level while considering obstacles such as walls.  To infer the best navigation strategy to reach a desired objective, the agent employs active inference and utilises the concept of Expected Free Energy (EFE). EFE is a measure of the agent's projected uncertainty or surprise about future states. By minimising EFE, the agent aims to reduce uncertainty and make accurate predictions about future outcomes, thus determining an optimal path to the objective \cite{nav_aif, curiosity_explitative}. This hierarchical active inference process, coupled with the integration of EFE, allows the agent to effectively explore new rooms at the highest level, navigate within the rooms at the mid-level, and execute actions seamlessly at the low level. 

The overall system can be represented as the formula equation \ref{eq:HAIF}. 


\begin{equation}
\begin{split}
P(\tilde{o}, \tilde{z}, \tilde{s}, \tilde{l}, \pi, \tilde{\pi_l}, \tilde{\pi_p}) & = P(\pi) \prod_T P(z_T, p_T^0 | l_T) P(l_T | \pi) P(\pi_l)  \\ & \prod_t P(s^t_0| z_T, p^T_t) P(p^T_t | \pi_l, p^T_0) P(\pi_p) \\ & \prod_\tau P(s^t_{\tau+1} | s^t_{\tau}, a^t_T) P(a^t_\tau | \pi_p) P(o^t_\tau, c^t_\tau | s^t_\tau) 
\end{split}
\label{eq:HAIF}
\end{equation}

Our model hyper-parameters are defined Appendix \ref{app:hyper_params}.  It is trained on a dataset of pixel observations collected by sampling random actions in a 3 by 3 rooms mini-grid environment as depicted in Fig\ref{img:3x3_env}. For more details on the models and training procedure we also refer to Appendix~\ref{app:our_model_train}.

\section{Results}

\subsection{Tasks oriented navigation}

Our testing primarily centre around assessing the model's capacity to execute specific functional tasks, including acquiring spatial maps through exploration in the presence of aliased and disjoint sensory input, transferring structural knowledge, and facilitating hierarchical planning. The agent is assigned two tasks: exploration and goal-reaching, both achieved without requiring re-training beyond learning familiar room structures.

\textbf{Baseline}.
To establish a baseline for the navigation tasks, we compare our method against: 
\begin{itemize}
    \item C-BET \cite{cbet}, an RL algorithm combining model-based planning with uncertainty estimation for efficient exploration decision-making.
    \item  Random Network Distillation (RND) \cite{RND}, integrates intrinsic curiosity-driven exploration to incentivise the agent's visitation of novel states, meant to foster a deeper understanding of the environment.
    \item Curiosity \cite{curiosity}, leverages information gain as an intrinsic reward signal, encouraging the agent to explore areas of uncertainty and novelty. 
    \item Count-based exploration \cite{count} uses a counting mechanism to track state visitations, guiding the agent toward less explored regions. 
    \item Dreamerv3 \cite{Dreamerv3} represents an advanced iteration of world models for RL, offering the potential to enhance navigation by predicting and simulating future trajectories for improved decision-making.
    \item A-star Algorithm (Oracle) \cite{Astar_dkjkstra}, is a path planning algorithm to which the full layout of the environment and its starting position is given to plan the ideal path to take between two points.
\end{itemize}

Each of these models propose various RL based exploration strategies for robotics navigation.
All baselines were trained and tested on the exact same environments. For each model training details, please refer to Appendix \ref{app:system_requirements} to \ref{app:observations}. 

The testing environments consist of connected rooms of increasing scale, ranging from 9 up to 20 rooms, each  room with a width of 4 tiles.

\subsubsection{Exploration}

We assess to what extent the hierarchical active inference model enables our agent to efficiently explore the environment. Without a preferred state leading the model toward an objective, the agent is purely driven by epistemic foraging, i.e. maximising information gain, effectively driving exploration \cite{AIF_learning}.

Our evaluation involves comparing the performance of various models, including our proposed hierarchical active inference model, C-BET, Count, Curiosity, RND models, and an Oracle. These models are tasked with exploring fully new environments with configurations ranging from 9 to 20 rooms. While the oracle possesses complete knowledge of the environment and its initial position, other models are equipped only with their top down view observations (and, in the case of some RL models, extrinsic rewards) -see Appendix\ref{app:observations} for more information about each model observation type-. The RL models are encouraged to explore until they locate a predefined goal (white tile); however, the reward associated with the white tile is muted to encourage continued exploration. Notably, the DreamerV3 model faces challenges in effective exploration due to its reliance on visual observations of the white tile for reward extraction. Consequently, an adapted environment without the white tile or a specific training would be necessary to employ DreamerV3 as an exploration-oriented agent in this study.

\begin{figure*}[htb!]
     \centering
     \begin{subfigure}[t]{0.21\textwidth}
         \includegraphics[width=\textwidth]{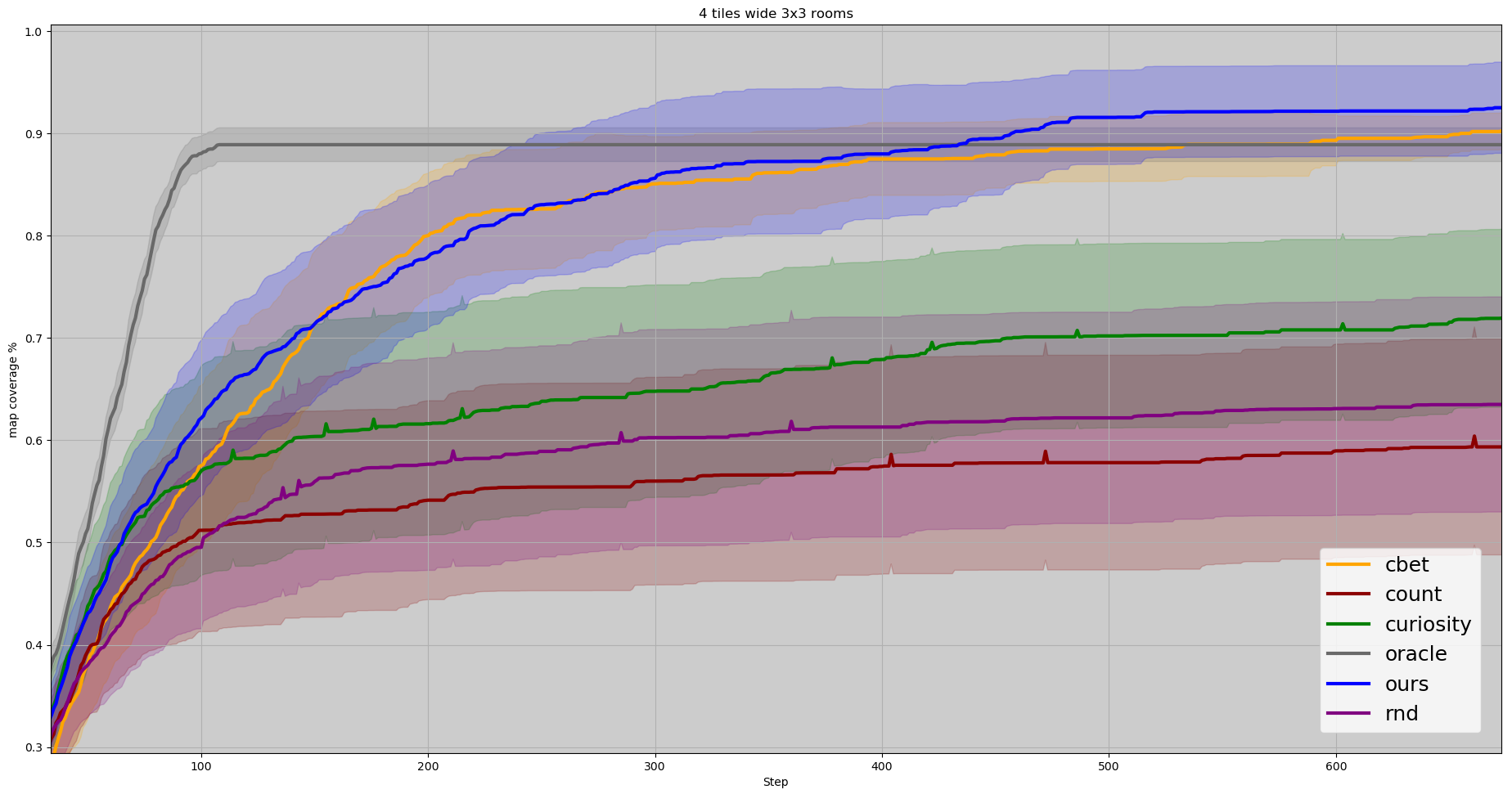}
        \caption{coverage over steps of all models in 3 by 3 rooms environments}
         \label{fig:4t_3x3_explo}
     \end{subfigure}
     \hfill
     \begin{subfigure}[t]{0.21\textwidth}
         \includegraphics[width=\textwidth]{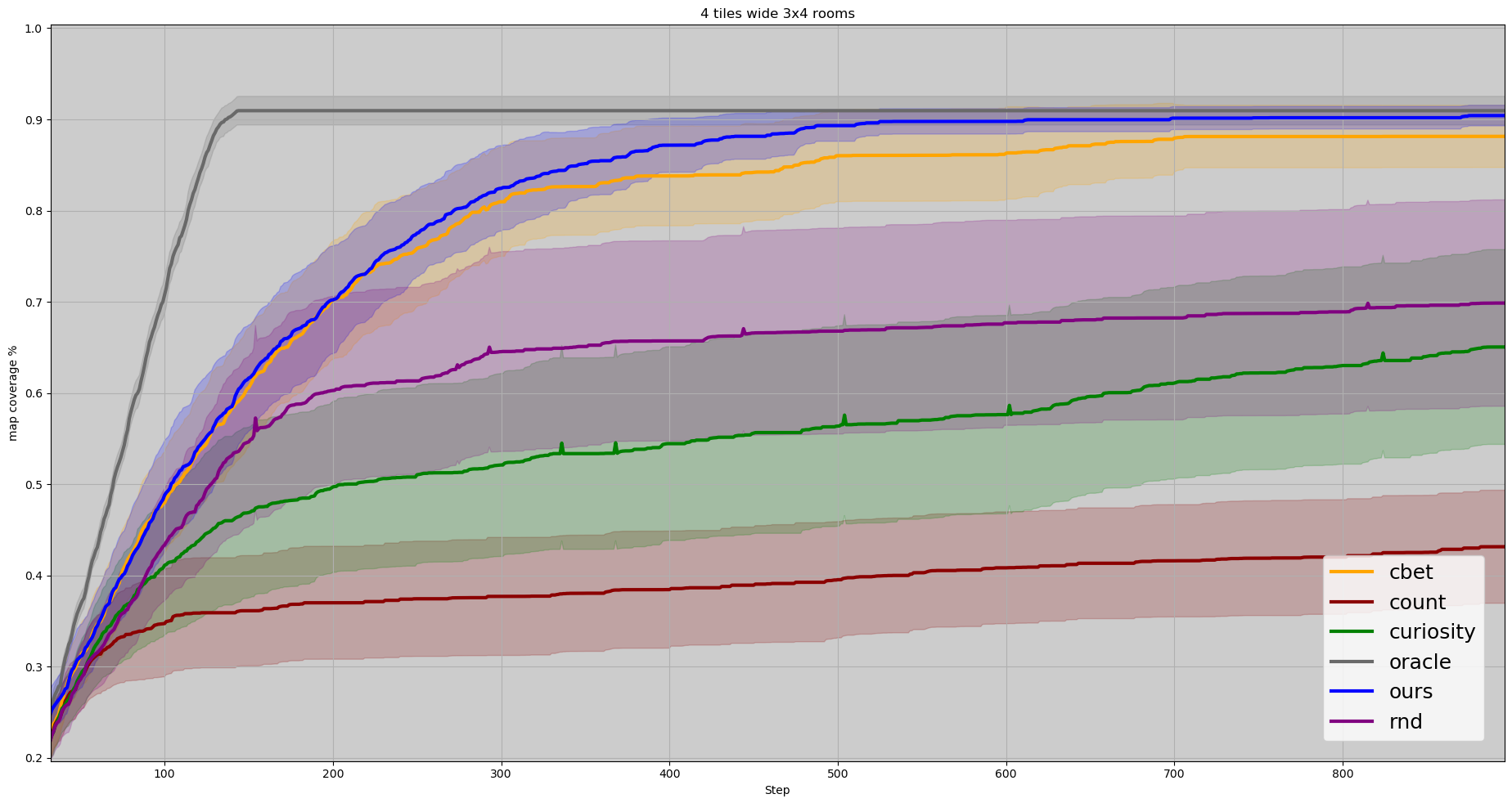}
         \label{fig:4t_3x4_explo}
         \caption{coverage over steps of all models in 3 by 4 rooms environments}
     \end{subfigure}
     \hfill
     \begin{subfigure}[t]{0.21\textwidth}
         \includegraphics[width=\textwidth]{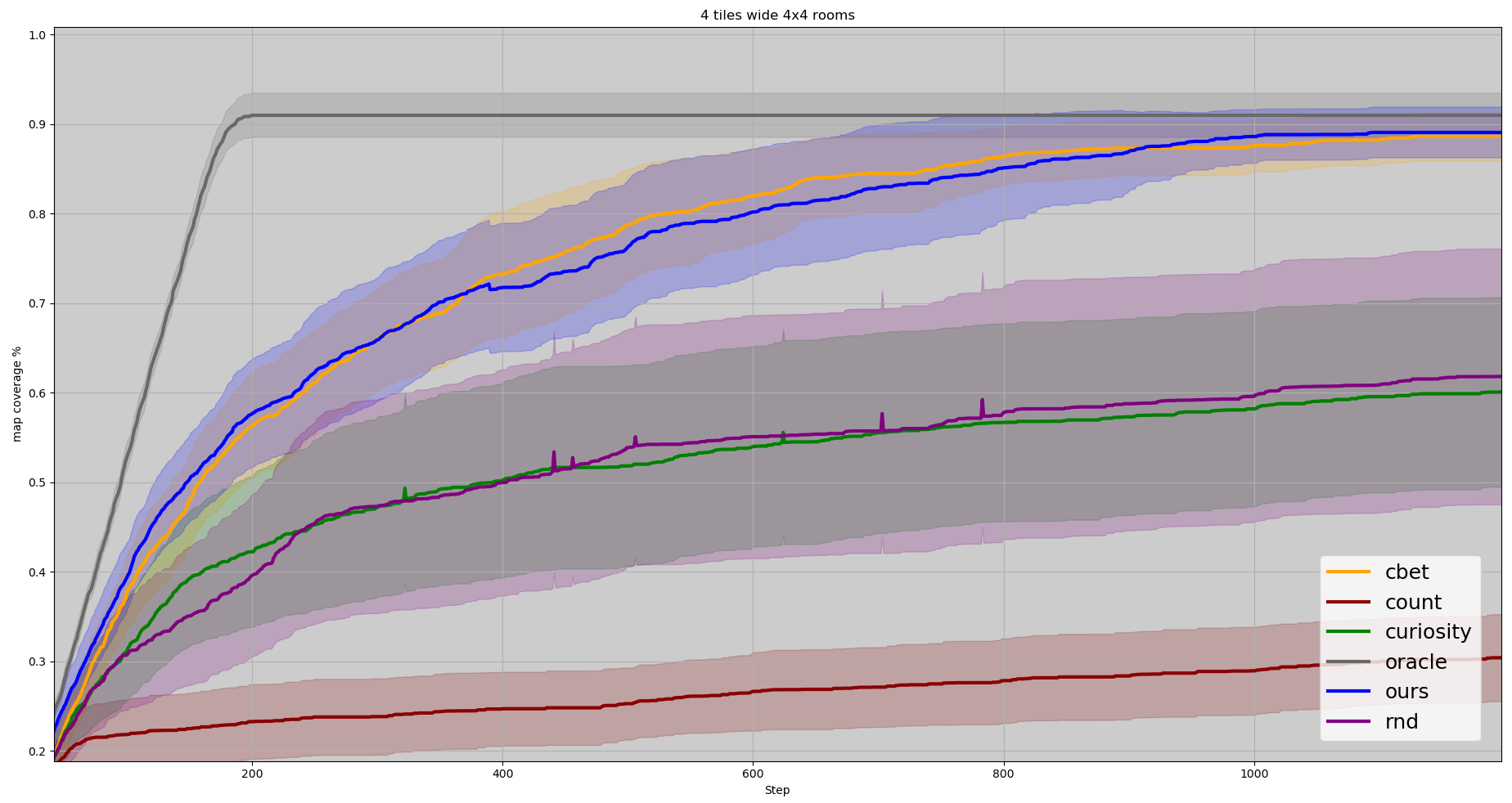}
        \caption{coverage over steps of all models in 4 by 4 rooms environments}
         \label{fig:4t_4x4_explo}
     \end{subfigure}
     \hfill
     \begin{subfigure}[t]{0.21\textwidth}
         \includegraphics[width=\textwidth]{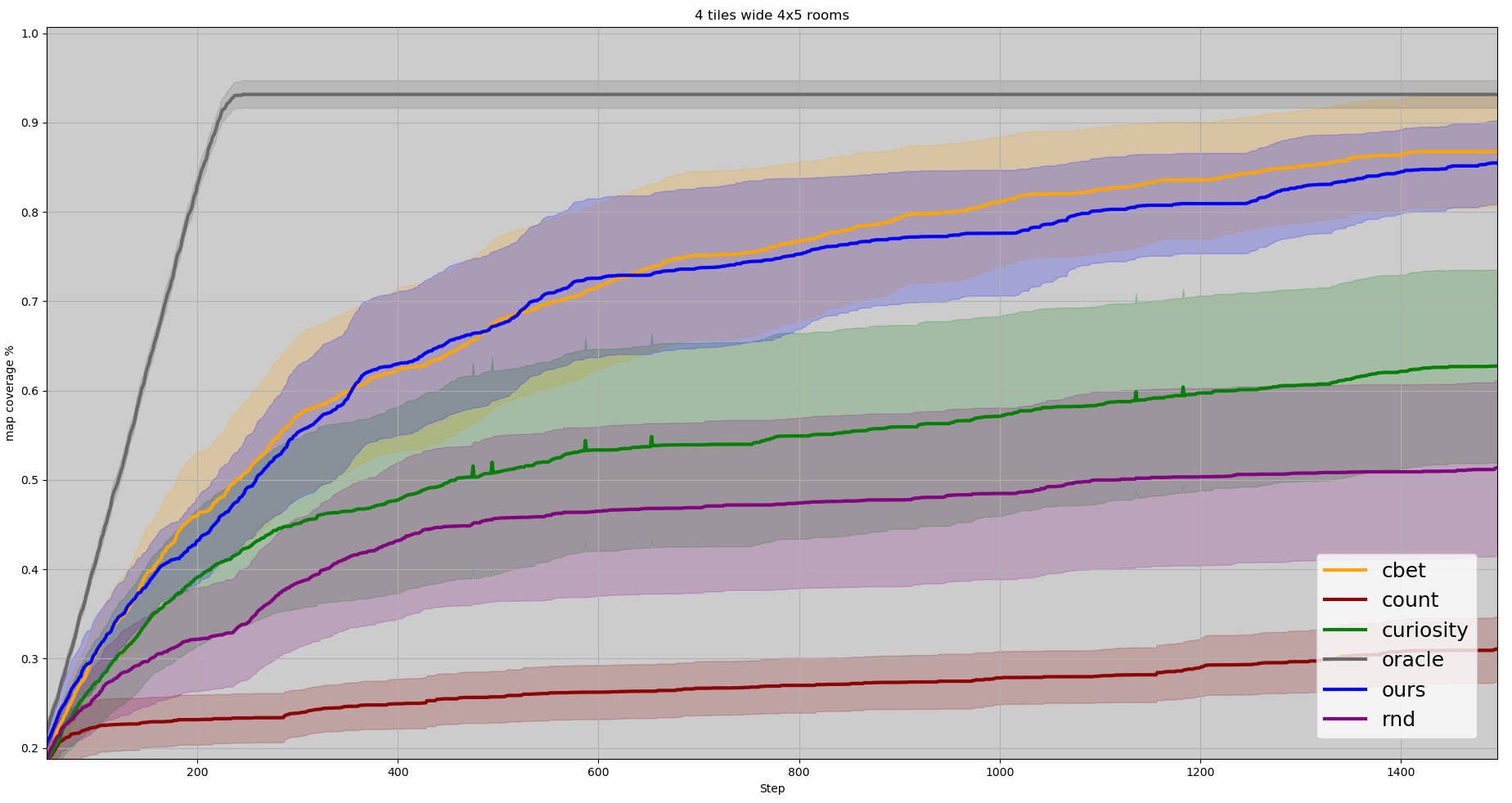}
         \label{fig:4t_4x5_explo}
         \caption{coverage over steps of all models in 4 by 5 rooms environments}
     \end{subfigure}
        \caption{The average exploration coverage across all test instances ($>$30runs) for each model computed within a certain scale of environment. Our model and C-Bet demonstrate similar performances in terms of speed and overall coverage, with our model exhibiting a narrower error margin in 3 by 4 rooms configuration, indicating consistent achievement of the specified coverage in most runs.}
        \label{fig:all_env_explo}
\end{figure*}

Across more than 30 runs by environment scale, our model demonstrates efficient exploration comparable to C-BET and notably outperforming other RL models in all tested environments, as depicted in Fig.\ref{fig:all_env_explo}. Moreover, the agent successfully achieves the desired level of exploration more frequently than any other model across all configurations, as demonstrated in Table \ref{tab:explo_success_rate}. For an exploration attempt to be considered successful, the agents must observe a minimum of 90\% of the observable environment. This criterion ensures that all rooms are observed at least once, without imposing a penalty on the models for not capturing every single corner. Since the agents cannot see through walls (see Appendix \ref{app:observations}), entering a room may result in missing the adjacent wall corners, but these corners hold limited importance for the agent's objective. As an unlikely example, missing all the corner tiles of each room results in 9\% of the environment not being observed (thus no matter the scale of the environment). 

\begin{table}[htb!]
\centering
\begin{tabular}{c|ccccc|}
\cline{2-6}
\begin{tabular}[c]{@{}c@{}}success \\ rate (\%)\end{tabular} & \multicolumn{5}{c|}{models} \\ \hline
\multicolumn{1}{|c|}{\begin{tabular}[c]{@{}c@{}}environment\\ configuration\end{tabular}} & \multicolumn{1}{c|}{ours} & \multicolumn{1}{c|}{C-BET} & \multicolumn{1}{c|}{RND} & \multicolumn{1}{c|}{curiosity} & count \\ \hline
\multicolumn{1}{|c|}{3x3 rooms} & \multicolumn{1}{c|}{\textbf{93}} & \multicolumn{1}{c|}{81} & \multicolumn{1}{c|}{16} & \multicolumn{1}{c|}{32} & 13 \\ \hline
\multicolumn{1}{|c|}{3x4 rooms} & \multicolumn{1}{c|}{\textbf{94}} & \multicolumn{1}{c|}{87} & \multicolumn{1}{c|}{16} & \multicolumn{1}{c|}{19} & 0 \\ \hline
\multicolumn{1}{|c|}{4x4 rooms} & \multicolumn{1}{c|}{\textbf{91}} & \multicolumn{1}{c|}{81} & \multicolumn{1}{c|}{26} & \multicolumn{1}{c|}{16} & 0 \\ \hline
\multicolumn{1}{|c|}{4x5 rooms} & \multicolumn{1}{c|}{\textbf{81}} & \multicolumn{1}{c|}{74} & \multicolumn{1}{c|}{7} & \multicolumn{1}{c|}{23} & 3 \\ \hline
\end{tabular}
\caption{The success rate of each model across all runs in each environment is defined as the percentage of runs where the exploration covers at least 90\% of the environment.}
\label{tab:explo_success_rate}
\end{table}

\subsubsection{Exploitative Behaviour}

\begin{figure}[htb!]
\vskip 0.05in
\begin{center}
\centerline{\includegraphics[width=9cm]{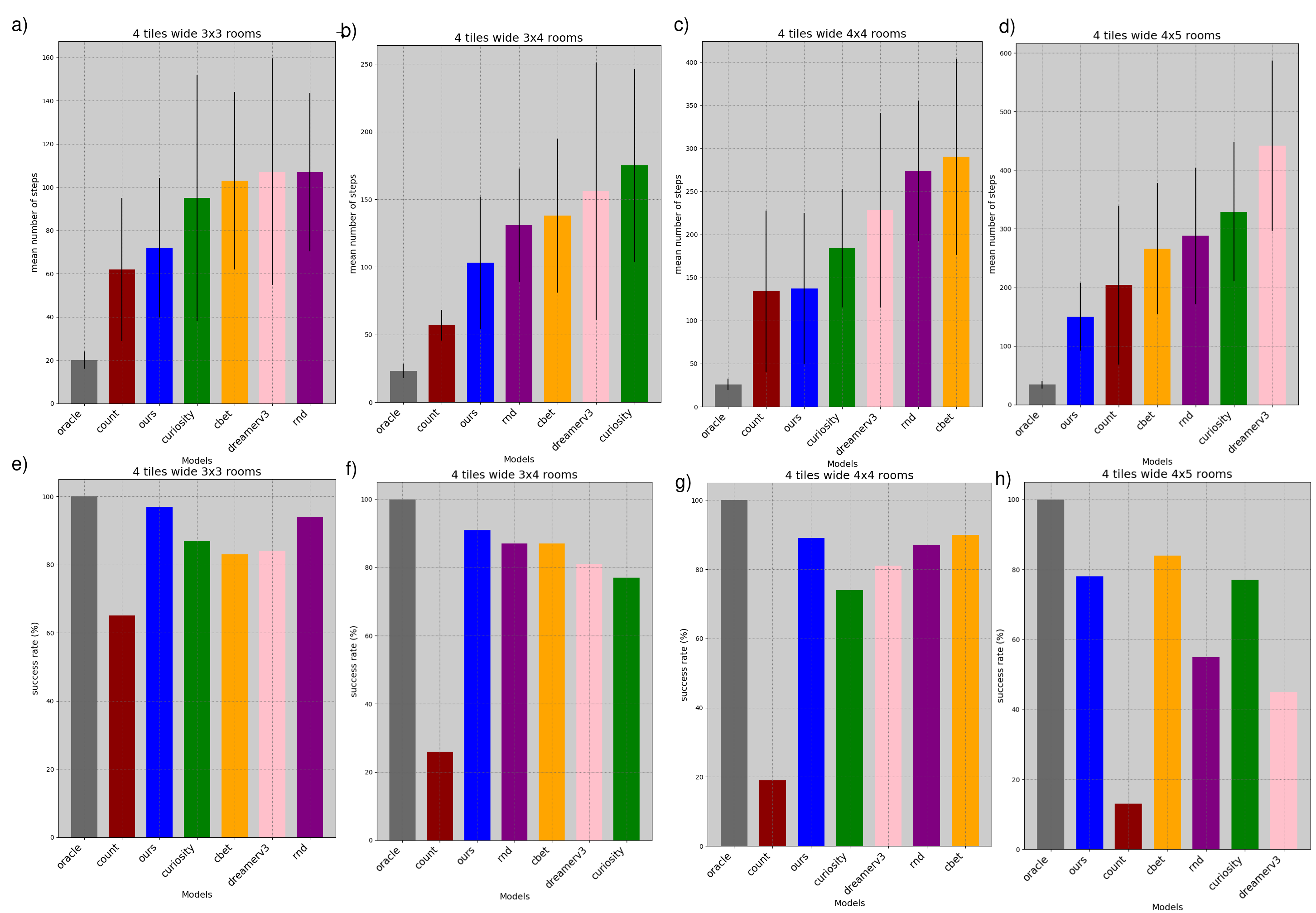}}
\caption{All models are assigned a goal seeking task in more than 30 environments varying in maze scale from 3 by 3 to 4 by 5 rooms presented in order from the 1$^{st}$ column to the last. The first row presents the mean number of steps each model take to successfully achieve the objective, with the black bars indicating a deviation of 15\% around the mean. The second row representing the success rate of each model in completing the task as a percentage. Across all environments, both C-BET and our model outperform the other RL models in terms of the mean number of steps required to reach the goal. Each model has its own colour: Grey for the Oracle, red for Count, blue for Ours, green for curiosity, orange for C-BET, pink for DreamerV3 and purple for RND.}
\label{img:standard_goal_reaching_task}
\end{center}
\vspace{-8mm}
\end{figure}

To evaluate the exploitative behaviour of the models, we configure all the models mentioned in the baseline to navigate to the single white tile within the environment. This is conducted across environments of escalating size, ranging from 9 to 20 rooms. Goal-directed behaviour is induced in our model by setting a preferred observation (i.e. the white tile) as typically done in active inference~\cite{AIF_learning, curiosity_explitative}. In the other RL models, an extrinsic and intrinsic reward is associated with this white tile, motivating the agents to explore until they reach this tile. The task is considered successful when the agent steps on the single white tile of the maze.
All the models, except the oracle, start without knowing their and the goal position in the environment, they therefore need to explore until they find the objective. Fig.\ref{img:standard_goal_reaching_task} displays the results of all the model in the goal seeking task in the diverse environments. Our model requires, in average, more steps than the Count model to reach the white tile in the 3 by 3 and 3 by 4 rooms configurations, however we can observe that count also has the lowest success rate. The Count model often fails when reaching the goal requires to cross several rooms. Overall our model reaches the white tile 89\% of the time over all environments (see Tab.\ref{tab:global_success_rate_goal}), Dreamerv3 is showing a poor performance because of over-fitting, not adapting well to new rooms configurations and white tile placement it has never seen during training. All models underwent training using the identical dataset detailed in Appendix \ref{app:dataset}, and the optimal models are selected for testing purposes. This observation suggests that Dreamerv3 might require a comparatively higher degree of human intervention to effectively operate within our environment compared to other models.

\begin{table}[htb!]
\centering
\begin{tabular}{|c|c|}
\hline
Models & \begin{tabular}[c]{@{}c@{}}success \\ rate (\%)\end{tabular} \\ \hline
Oracle & 100\% \\ \hline
Ours & 89\% \\ \hline
C-BET & 86\% \\ \hline
RND & 81\% \\ \hline
Curiosity & 79\% \\ \hline
DreamerV3 & 72\% \\ \hline
Count & 31\% \\ \hline

\end{tabular}
\caption{The success rate of each model across all environments and runs.}
\label{tab:global_success_rate_goal}
\end{table}

\subsection{Qualitative results}

\begin{figure}[b]
\vspace{-5mm}
\begin{center}
\centerline{\includegraphics[width=8cm]{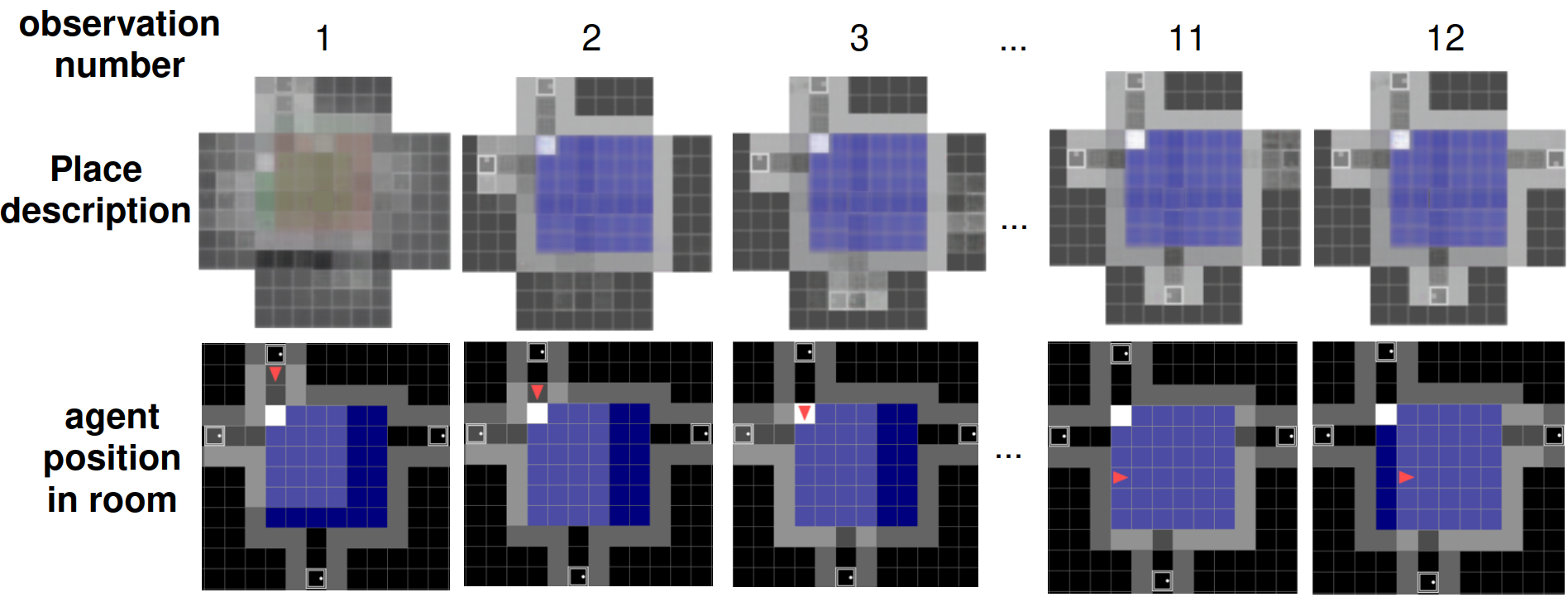}}
\caption{Evolution of the place representation in a room as new observations are provided by the moving agent (red triangle). The model is able to correctly reconstruct the structure of the room as observations are collected.}
\label{img:room_generation}
\end{center}
\vskip -0.1in
\end{figure}

Figure \ref{img:room_generation} illustrates the inference process of place descriptions. Within approximately three steps, the main features of the environment are captured and form a reasonably accurate picture of the seen observations. When encountering a new aisle for the first time at step 11, the model is able to adapt and generate a well-imagined representation. Each observation corresponds to the red agent's clear field of view, as depicted in the agent position row of the figure (more details about the observations Appendix\ref{app:observations}).

\begin{figure}[htb!]
     \centering
      \begin{subfigure}[t]{0.5\textwidth}
         \centering
         \includegraphics[width=6.7cm]{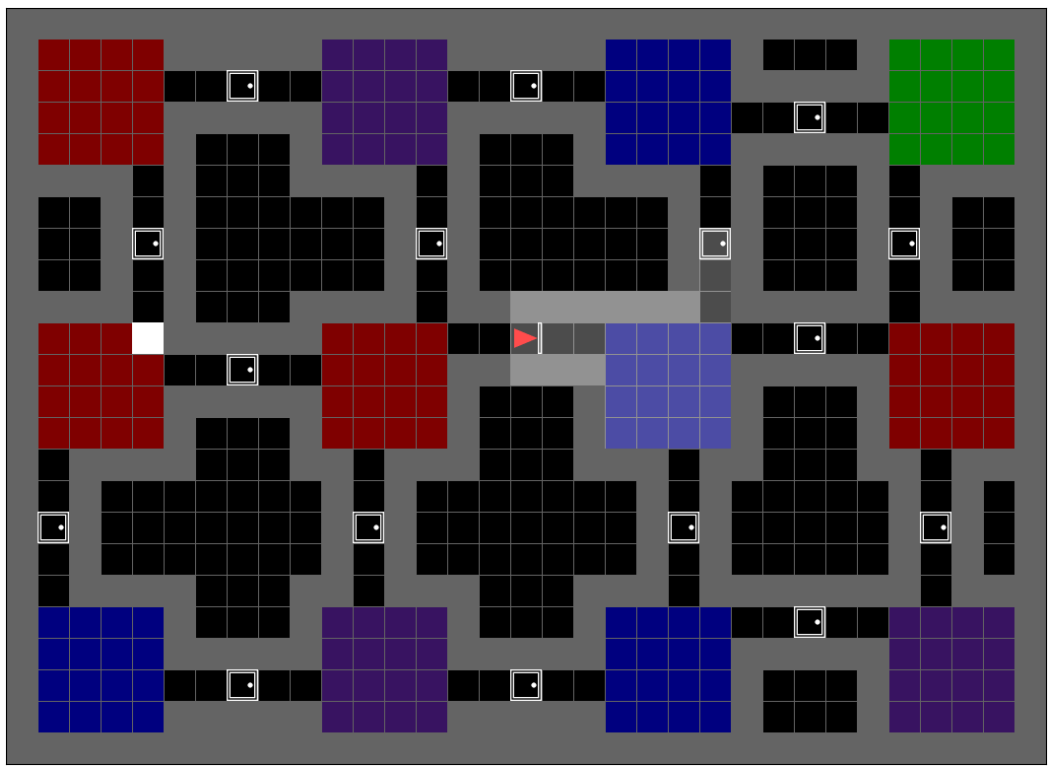}
        \caption{A 3 by 4 room-maze seen from above.}
         \label{fig:real_map}
     \end{subfigure}
     \hfill
     \begin{subfigure}[t]{0.5\textwidth}
         \centering
         \includegraphics[width=6.5cm]{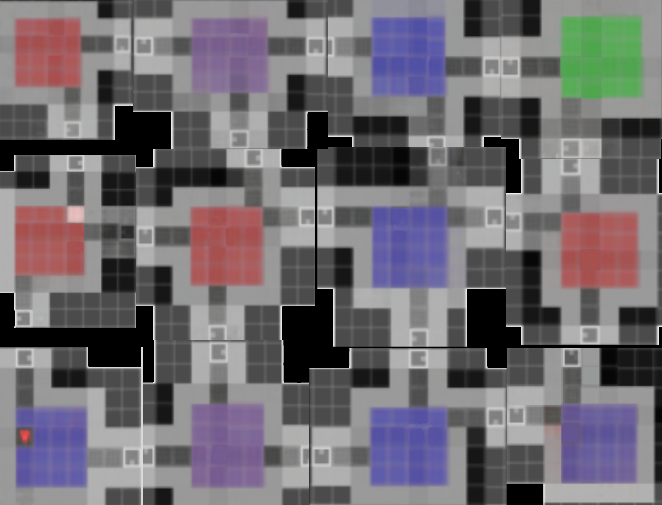}
        \caption{A map of a 3 by 4 environment and rooms representation given by the cognitive map. We can see that the estimated map is loyal to the real map aside from some non precisely defined aisles and the bottom right room having a slightly wrong aisle position.}
         \label{fig:our_memory_map}
     \end{subfigure}
     \caption{a) displays the real map while b) is a composition of a cognitive map room's representations. Overall, the rooms are recognisable when compared to the real map.}
     \label{fig:real_vs_cognitive_maps}
\end{figure}

Figure \ref{fig:real_vs_cognitive_maps} provides a direct comparison between the accuracy of the cognitive map's room reconstruction and the corresponding physical environment. This comparison reveals that the estimated map closely aligns with the actual map, with only minor discrepancies observed in some blurry passageways and a slight misplacement of the aisle in the bottom right room. This shows how important global position estimation is as the cognitive map uses believed location to distinguish  between two identical rooms (purple rooms in the second column). This alignment between real and imagined map underscores the fidelity of our model's internal representation in capturing the structural layout of the environment.  A supplementary demonstration showing the model's ability for place recognition can be found in Appendix\ref{Appendix_results} Fig.\ref{img:proba_rooms}.

Additional quantitative and qualitative findings are available in Appendix \ref{Appendix_results}, illustrating the agent's capacity to generalise and adeptly reconstruct more expansive environments beyond its training exposure. The appendix also presents an instance of navigation within a maze characterised by aliasing, showcasing the model's ability in constructing and selecting pertinent place representations. Furthermore, we demonstrate the proficiency of our hierarchical model in making precise predictions over extended time-frames, encompassing transitions across various rooms where, in contrast, recurrent state space models often encounter challenges when tasked with predicting across room boundaries. Lastly, we highlight the computational efficiency of our system in comparison to the conventional RL models employed in this study (Appendix \ref{app:system_requirements} and \ref{Appendix_results} Tab.\ref{tab:testing_requirements} ).

\section{Discussion}

The discussion section of this paper aims to provide a comprehensive analysis of the proposed hierarchical active inference model for autonomous navigation, considering its strengths and limitations. We outline the key contributions of our work and discuss the potential future works.

\textbf{Hierarchical Active Inference Model}. Our proposal introduces a three-layered hierarchical active inference model :
\begin{itemize}
    \item     The cognitive map unifies spatial representation and memorises location characteristics.
    \item     The allocentric model creates chunked spatial representations.
    \item     The egocentric model assesses policy plausibility, considering dynamic limitations.
\end{itemize}
These layers collaborate at different time scales: the high level oversees the whole environment through locations, the allocentric model refines place representations as it change position, and the egocentric model imagines action consequences.

\textbf{Low Computational Demands}. Our hierarchical active inference model has a low computational demands regardless of the environment's scale. This efficiency is particularly valuable as environments scale up, making our approach a potential solution for real-world applications.

\textbf{Scalability}. Our model efficiently learn spatial layouts and their connectivity. There exists the potential for our approach to adapt to novel scenarios by incorporating diverse environments into its learning process, thus expanding allocentric representations. Furthermore, the possibility of introducing additional higher layers could facilitate greater abstraction, transitioning from room-level learning to broader structural insights.

\textbf{Task Agnostic}. The system doesn't require task-specific training, promoting adaptability to diverse navigational scenarios. It learns environment structures and generalises to new scenarios, demonstrating applicability to various objectives.

\textbf{Visual based navigation}. Leveraging visual cues should enhance our model's real-world applicability.

\textbf{Aliasing Resistant}. We show resistance to aliases, distinguishing between identical places and thus ideally supporting robust navigation.

While our approach offers several advantages, it is also important to acknowledge its limitations:

\textbf{Environment Adaptation} Our model requires adaptation to fully new environments for optimal performance. Training the allocentric model on room-specific data restricts navigation to familiar settings. To mitigate this, and generalise to arbitrary environments, we could consider splitting the data by unsupervised clustering~\cite{Self-labelling}, or by using the model's prediction error to chunk the data into separate spaces~\cite{verbelen2022chunking}.

\textbf{Recognition of Changed Environments} Our proposal might struggle to detect environmental changes like altered tile colors, although this may not significantly impact navigation performance as a new place will replace or be added with the previous one in the cognitive map, it remains an area for improvement.



Our comprehensive assessment, both quantitative and qualitative, underscores the adaptability and resilience of our approach, which fares well in explorative and exploitative tasks compared to C-BET \cite{cbet}, Random Network Distillation (RND) \cite{RND}, Curiosity \cite{curiosity} or Count \cite{count}, Dreamerv3 \cite{Dreamerv3}. 

In conclusion, while our navigation demonstrates promising exploration and goal-reaching abilities in a minigrid \cite{gym_minigrid} environment by learning structures and leveraging visual cues, future research could focus on enhancing adaptation to new environments, handling changes in familiar ones, and adding comprehension to the cognitive map. Scalability and extension to complex dynamic scenarios are also potential avenues for exploration. While our model demonstrates promising performance in a mini-grid environment, its application to more realistic scenarios, such as the Memory maze  \cite{memory_maze} or Habitat~\cite{habitat}, could lead to more practical implementations. Further improvements could also involve learning a prior over topological map structures, promoting informed exploration and imagination of shortcuts~\cite{shortcuts}.

\section*{Acknowledgment}
This research received funding from the Flemish Government under the “Onder-
zoeksprogramma Artificiële Intelligentie (AI) Vlaanderen” programme.


\bibliographystyle{IEEEtran}
\bibliography{IEEEabrv,main}

\clearpage
\onecolumn
\appendix


The appendixes are composed of two parts, the training specifics detailed from the models required computational consideration Appendix \ref{app:system_requirements}, then describing the dataset App.\ref{app:dataset} then describing for each model the hyperparameters used if any different from their source paper and the observations used for their model App.\ref{app:hyper_params} and each model observation type App.\ref{app:observations}. Moreover Appendix \ref{app:our_model_train} details our training more in depth with the loss functions used for optimisation. Finally App.\ref{Appendix_results} shows all the additional tests results from quantitative assessment of the generalisation ability to represent places to the models testing computational requirements.

\subsection{Training system requirements}

\label{app:system_requirements}
\begin{table}[!htb]
\centering
\begin{tabular}{|c|c|c|c|c|c|c|}
\hline
model & \begin{tabular}[c]{@{}c@{}}Training \\ time (h)\end{tabular} & n$^{\circ}$ CPU & n$^{\circ}$ GPU & \begin{tabular}[c]{@{}c@{}}used \\ RAM (G)\end{tabular} & \begin{tabular}[c]{@{}c@{}}used \\ Memory (G)\end{tabular} & GPU type \\ \hline
\begin{tabular}[c]{@{}c@{}}Ours\\ egocentric\end{tabular} & 32 & 4 & 1 & \_ & 12 & GTX 980 \\ \hline
\begin{tabular}[c]{@{}c@{}}Ours \\ allocentric\end{tabular} & 95 & 2 & 1 & 2.5 & 20 & GTX 1080 \\ \hline
Dreamerv3 & 411 & 5 & 2 & 10 & 30 & GTX 1080 Ti \\ \hline
C-BET & 232 & 10 & 1 & 2.6 & 32 & GTX 980 \\ \hline
RND & 117 & 6 & 1 & 2.7 & 10 & GTX 980 \\ \hline
Curiosity & 90 & 6 & 1 & 3 & 10 & GTX 980 \\ \hline
Count & 141 & 6 & 1 & 2.7 & 11 & GTX 980 \\ \hline
\end{tabular}
\caption{Comprehensive insights into the training specifics of all models are provided, encompassing their respective training duration until reaching their finalised versions. Additionally, details regarding the computational resources employed are presented including the maximum RAM and memory allocation required for each model. Notably, our model underwent training for both egocentric and allocentric components, executed in parallel using the same dataset, their division into distinct sets realised for practical reasons. Unfortunately, the information pertaining to the RAM utilisation by the egocentric model is unavailable.}
\label{tab:train_info}
\end{table}

\subsection{Training Dataset}
\label{app:dataset}
Uniformity in training conditions was achieved by conducting training sessions for all models within identical environments, facilitated by the consistent application of a shared seed to generate these environments. 
The model is trained on a mini-grid environment consisting of 3 by 3 squared rooms of 4 to 7 tiles wide connected by aisles of fixed length randomly placed, separated by a closed door in the middle. Each room is assigned a colour at random from a set of four: red, green, blue, and purple. In addition, white tiles may be present at random positions in the map.
The agent has a top view of the environment covering a windows of 7 by 7 tiles, including it's own occupied tile. It cannot see behind itself, nor through walls or closed doors.

\subsection{Hyper-Parameters}
\label{app:hyper_params}

All the adversarial models were trained upon pre-set hyper-parameters, with C-BET, Count, Curiosity and RND upon \cite{cbet} described parameters. 
And DreamerV3 upon \cite{Dreamerv3} proposed work, however the behaviour was modified from the original, setting an Exploring task behaviour and a Greedy exploration behaviour as the original configuration was over-fitting in our scenarios.

Our model was trained using the hyper-parameters in Tab.\ref{table:gqn params} for the allocentric model and Tab.\ref{table:oz params} for the egocentric model 

\begin{table}[!h]
\centering
\scalebox{0.7}{
\begin{tabular}{llcc}
                                               & \textbf{Layer} & \multicolumn{1}{l}{\textbf{Neurons/Filters}} & \multicolumn{1}{l}{\textbf{Stride}} \\ \hline
\multirow{1}{*}{PositionalEncoder}             & Linear         & 9                                           &                                     \\ \hline

\multirow{3}{*}{Posterior}                 & Convolutional  & 16                                           & 1 // (kernel:1)                      \\
\multicolumn{1}{c}{}                           & Convolutional  & 32                                           & 2                                   \\
\multicolumn{1}{c}{}                           & Convolutional  & 64                                           & 2                                   \\
\multicolumn{1}{c}{}                           & Convolutional  & 128                                           & 2                                   \\
                                               & Linear         & 2*32                                         &                                     \\ \hline
\multicolumn{1}{c}{\multirow{8}{*}{Likelihood}} & Concatenation  &                                           &                                    \\
                                               & Linear         & 256*4*4                                      &                                     \\
                                               & Upsample       &                                              &                                    \\
                                               & Convolutional  & 128                                          & 1                                   \\
                                               & Upsample       &                                              &                                     \\
                                               & Convolutional  & 64                                           & 1                                   \\
                                               & Upsample       &                                              &                                     \\
                                               & Convolutional  & 32                                           & 1                                   \\
                                               & Upsample       &                                              &                                     \\
                                               & Convolutional  & 3                                            & 1                                   \\ \hline       
\end{tabular}
}
\caption{allocentric model parameters}
\label{table:gqn params}
\end{table}

\begin{table}[!h]
\centering
\scalebox{0.7}{
\begin{tabular}{llcc}
                                               & \textbf{Layer} & \multicolumn{1}{l}{\textbf{Neurons/Filters}} & \multicolumn{1}{l}{\textbf{Stride}} \\ \hline
\multirow{3}{*}{Prior}                         & Concatenation  &                                              &                                     \\
                                               & LSTM           & 256                                          &                                     \\
                                               & Linear         & 2*32                                         &                                     \\ \hline
\multicolumn{1}{c}{\multirow{8}{*}{Posterior}} & Convolutional  & 8                                           & 2                                   \\
\multicolumn{1}{c}{}                           & Convolutional  & 16                                           & 2                                   \\
\multicolumn{1}{c}{}                           & Convolutional  & 32                                           & 2                                   \\
\multicolumn{1}{c}{}                           & Concatenation  &                                              &                                     \\
\multicolumn{1}{c}{}                           & Linear         & 256                                          &                                     \\
\multicolumn{1}{c}{}                           & Linear         & 64                                         &                                     \\ \hline

\multirow{4}{*}{Image\_Likelihood}             & Linear         & 256                                          &                                     \\
                                               & Linear         & 32*7*7                                      &                                     \\
                                               & Upsample       &                                              &                                    \\
                                               & Convolutional  & 16                                          & 1                                   \\
                                               & Upsample       &                                              &                                     \\
                                               & Convolutional  & 8                                           & 1                                   \\
                                               & Upsample       &                                              &                                     \\
                                               & Convolutional  & 3                                           & 1                                   \\  \hline
\multirow{3}{*}{Collision\_Likelihood}         & Linear  &       16                                       &                                     \\
                                               & Linear           & 8                                          &                                     \\
                                               & Linear         & 1                                         &     

                                               \\ \hline
                                            
\end{tabular}
}
\caption{egocentric model parameters}
\label{table:oz params}
\end{table}

\subsection{Models observations}
\label{app:observations}
\begin{figure}[htb!]
\centering
\includegraphics[width=9cm]{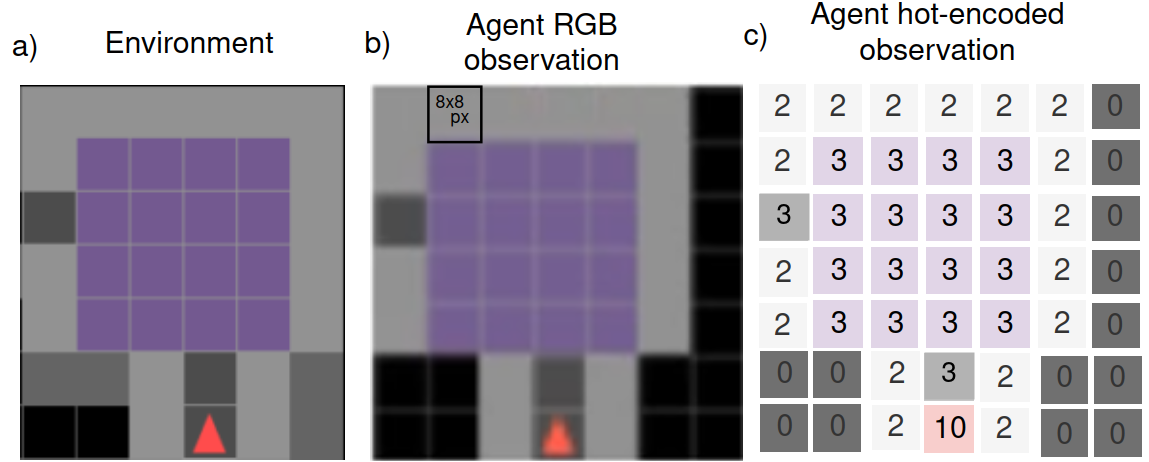}
\caption{a) cropped top down view of the environment, b) the RGB view of the agent, each tile of the environment are composed of 8 by 8 pixels, generating a 56 by 56 total image. c) the equivalent hot-encoded view as a matrix, the numbers and colours are only relevant for the example.}
\label{app_img:ob}
\end{figure}

All models use the agent's top down vision of the agent, consisting in 7 by 7 tiles with the agent placed at the bottom centre of the image, as shown in \ref{app_img:ob}. Our model and DreamerV3 use an RGB pixel rendering of shape 3x56x56. The observation the agent interprets is a . while C-BET, Count, Curiosity and RND use a flat hot-encoded view of the environment as well as an extrinsic reward when passing over the single white tile in the environment. We can point out that the agent can't see through walls in RGB image, we can see in Fig.\ref{app_img:ob} a) the environment and the agent's field of view represented by lighter colours.  Fig.\ref{app_img:ob} b) shows the actual observation seen by the agent.

The number of actions cbet could take is greatly reduced compared to the original work, limiting to actions such as forward, left, right and stand-by.

All the RL models have a sparse reward system, with an extrinsic reward generated only when passing on the white tile disposed in the environment. Our model doesn't require rewards, as such the goal we desire to set during the testing could be any kind of observation.

\subsection{Our Model training}
\label{app:our_model_train}

In order to effectively train this hierarchical model, the two lower level models are considered independent and trained in parallel
To optimise the two ego-allocentric neural network models we first obtain a dataset of sequences of action-observation pairs by interacting with the environment. This can for example be obtained using a random policy or by human demonstrations.

The egocentric and allocentric models are considered independent to optimise training. While they are trained over the same data, the allocentric model has sequences chunked up to maximum the transition between two rooms, so that each sequence encompass only a room each time.
Thus the allocentric model received as training data of 1000 sequences of 40steps per room size (4 to 7 tiles width) while the egocentric received 100 sequences of 400 steps per room size, each full sequence then cut in sub-sequences of 20 steps. All the training took place in 3x3 rooms maze where the agent could start the sequence randomly from any door (or near door) position. The training was realised on 100 environments per room size.
Both neural networks can be trained in parallel end-to-end on this dataset using stochastic gradient descent by minimising the free energy loss function.\\
For the egocentric model:

\begin{equation} 
\mathcal{L} = \sum_{t=1}^T {D_{KL}[p_\phi(s_t|s_{t-1}, a_{t-1}, o_t)||p_\theta(s_t|s_{t-1},a_{t-1})]} - log[ p_\xi(o_t|s_t)]
\end{equation}

The egocentric model was trained by minimising, in one part, the difference between the expected belief state given the couple action, previous history and the estimated posterior obtained given the action, observation and updated history. And in a second part minimising the difference between the reconstructed observation and the input observation \cite{robot_nav_oz}.

While for the allocentric model:
\begin{equation} 
\mathcal{L} = \sum_{t=0}^T {D_{KL}[Q\phi(z|o_t, p_t)||\mathcal{N}(0,1)]} + ||\hat{o}_t - o_t||^{2}
\end{equation}

The approximate posterior Q is being modelled by the factorisation of the posteriors after each observation.
The belief over z can then be acquired by multiplying the posterior beliefs over z for every observation. We learn an encoder neural network with parameters $\phi$ to learn the posterior state z given
a single observation and pose pair $(o_k, s_k)$.

And the Likelihood being optimised by an MSE given the real observation $o_k$ and the predicted observation $\hat{o}_k $ \cite{GQN_Toon}, for each room the posterior is built on random sequence length varying from 15 observations up to the whole sequence of 40 steps. In order to obtain a position, the action of the agent is integrated into the next position.  
Both models use an Adam optimisation \cite{adam_optimiser}.

While the cognitive map has been adapted for navigation in this type of mini-grid world \cite{gym_minigrid}, it is thought to be re-scaled or adapted to other environments.
\\

\subsection{Supplementary results}
\label{Appendix_results}
The following appendix contributes to the work by shedding additional lights on the ability of the model to solve rooms aliasing and its ability to generalise.

Fig3 shows the agent consistently achieving a stable place description within about three observations in room sizes that were part of its training. Interestingly, the agent also demonstrates the ability to accurately reconstruct larger rooms, even though it did not encounter such sizes during training. In particular, stable place descriptions for 8-tile wide rooms are achieved in approximately five steps. This showcases the agent’s allocentric model generalisation abilities beyond the limits of its training. The experiment was conducted over 125 runs in 25 environments with the agent tasked to predict observations from unvisited poses after each new motion. Fig4 demonstrate the significance of the Mean Squared Error (MSE) value by displaying examples of predicted observations and their corresponding MSE values. It can be observed that under an MSE of 0.5, the predictions of the observations are visually quite accurate.

\begin{figure}[hbt!]
  \centering
  \begin{minipage}[b]{0.4\textwidth}
    \includegraphics[width=\textwidth]{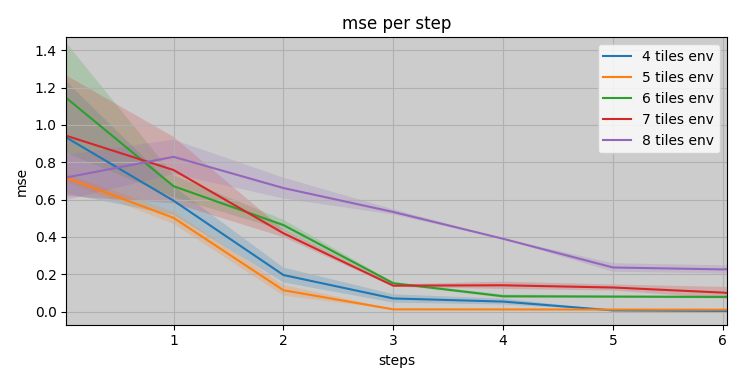}
    \caption{Prediction error of unvisited positions over min 5 tests per 5 environments by room size starting from step 0 where the models has no observation.}
    \label{img:MSE_envs}
  \end{minipage}
  \hfill
  \begin{minipage}[b]{0.5\textwidth}
    \includegraphics[width=\textwidth]{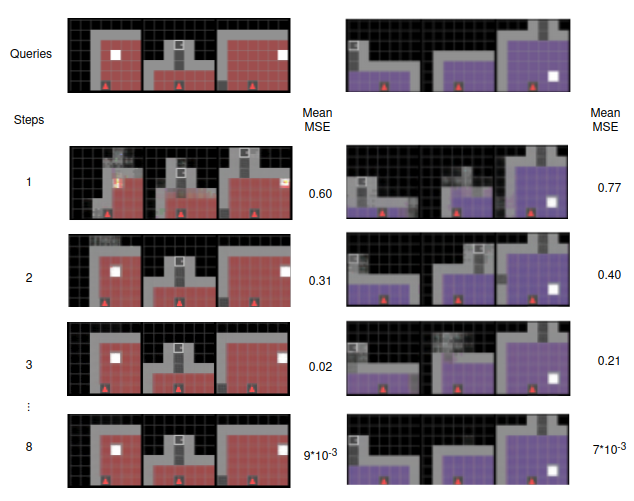}
    \caption{Observation queries, sampled prediction over unvisited position and mean MSE over 5 samples of predictions}
    \label{img:MSE_eg}
  \end{minipage}
\end{figure}

\begin{figure}[hbt!]
\centering
\includegraphics[width=16cm]{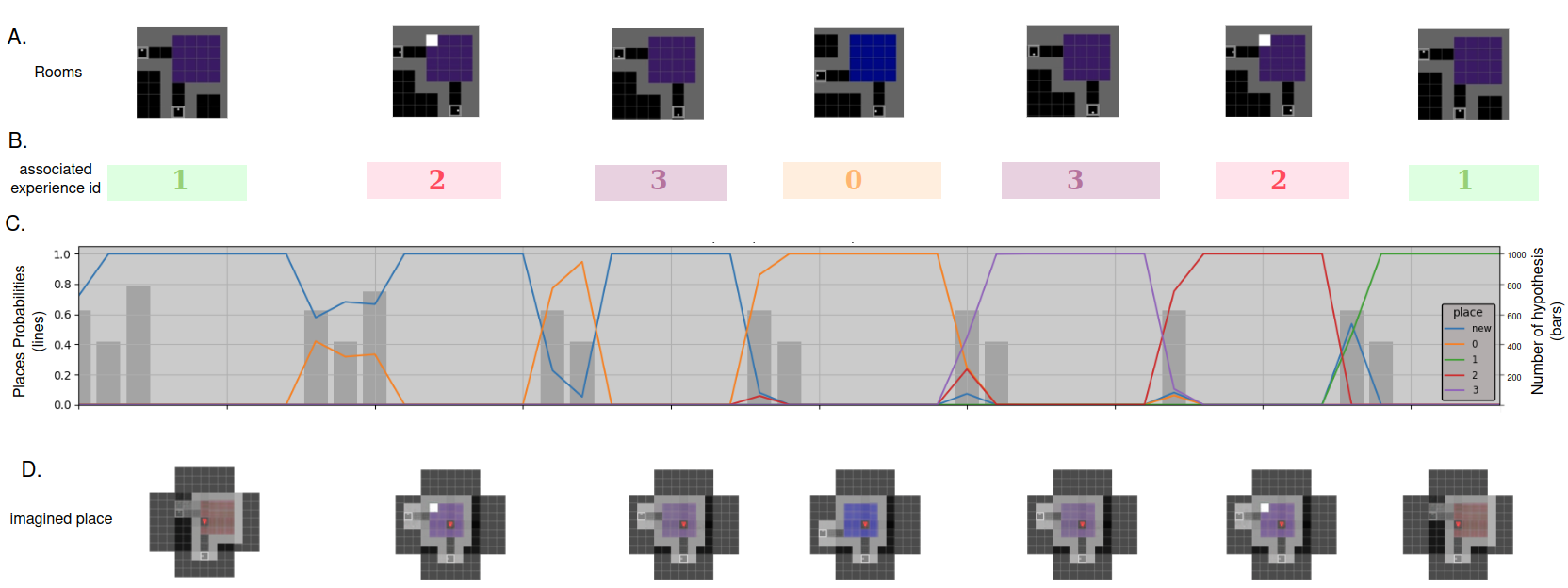}
\caption{Navigation samples of the agent looping clockwise and anti-clockwise (resulting in entry through different doors) in a new 2 by 2 room environment. Clockwise navigation corresponds to entirely new exploration, generating new places (as seen with the blue lines in C. corresponding to a new room generation), while the anti-clockwise loop traverses previously explored places. A. shows a new world comprising four visually similar rooms (identical in colour, shape, or both). B. illustrates the model's association of each room with a distinct experience ID  C.  C depicts the probability of a new place being created (in blue, representing the most probable place among all possibilities) or an existing place being deemed the most likely to explain the environment. The grey bars indicate the number of new places considered concurrently, with the count of simultaneous hypotheses displayed on the right side of the plot. D. showcases the imagined place generated for each experience ID. Experience 1 is not entirely accurate, yet it suffices to distinguish it from other rooms given actual observations of it.}
\label{img:proba_rooms}
\end{figure}

In order to navigate autonomously, an agent must be capable of self-localisation and position correction based on visual information and its internal beliefs about the environment. We perform navigation in a highly aliased mini-grid maze consisting of four interconnected rooms. These rooms shared similarities in colour, configuration, or both colour and configuration, differing only by a single white tile. These four rooms are depicted in Figure \ref{img:proba_rooms} A. 
The complete Figure \ref{img:proba_rooms} demonstrates the agent's exploration of the rooms and its ability to differentiate between them without becoming confused, even when retracing its path from the starting room by entering rooms through different aisles than before.

effectively, when the agent identifies a new place, it creates a fresh experience by incorporating its location. Figure \ref{img:proba_rooms} B. displays each newly generated experience with a distinct ID and colour.
has entered a new place or returned to a familiar one, the agent considers the probability of each place to explain the current observations, as depicted in Figure \ref{img:proba_rooms} C. The bars illustrate the number of hypotheses considered at each step, and the lines represent the probability of the place being either a new or a previously visited one. The colours of the lines correspond to the colours attributed to the experiences in Figure \ref{img:proba_rooms} B, with blue lines representing new unidentified places.
Figure \ref{img:proba_rooms} D. displays the internal representation of the places the agent uses. It is evident that the rooms are accurately imagined, and even a doubt in an aisle position in experience 1 is not sufficient to confuse the agent.

In this context, the agent demonstrates the capability to navigate effectively and distinguish between rooms in a novel, highly aliased environment. The agent's aptitude to identify previously visited rooms even upon entering from a new doorway underscores its capacity to retain a spatial memory of the environment.

\begin{figure}[hbt!]
\centering
\includegraphics[width=16cm]{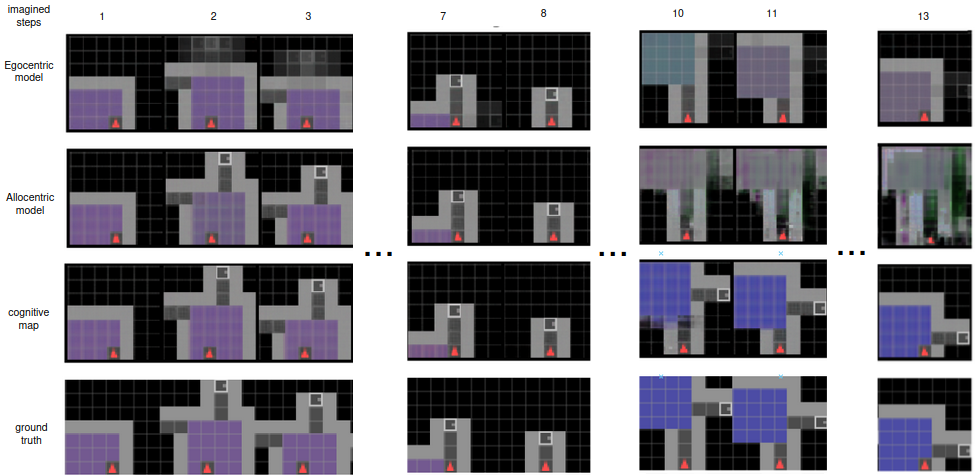}
\caption{A trajectory leading toward a previously visited room is imagined by each model's layer.
The egocentric model, characterised by its short-term memory, gradually loses information as time progresses. This is evident from step 2 onward, where the front aisle is no longer represented after the agent makes a few turns without visual input. In contrast, the allocentric model maintains the place description over time but encounters difficulty once it moves beyond the current place it occupies. The cognitive map, possessing knowledge of the connections between locations, accurately deduces the expected place behind the door, resulting in a prediction remarkably similar to the ground truth.}
\label{img:traj_over_layers}
\end{figure}

Furthermore, our hierarchical model also facilitates accurate predictions over extended timescales that span different rooms. In contrast, recurrent state space models commonly struggle when tasked with predicting across room boundaries. Fig\ref{img:traj_over_layers} illustrates the prediction capabilities of each layer over a prolonged imagined trajectory within a familiar environment. The figure showcases the predictions that each layer of the model would make as we project the imagination into the future, up to the point of transitioning to a new room and beyond.
The first row demonstrates how the egocentric model gradually loses the spatial layout information over time, making it more suitable for short-term planning. The second row highlights the model's limitation to a single place in the environment, failing to recognise the subsequent room as the same place. Lastly, in the third row, the cognitive map's imagined trajectory accounts for the agent's location and is capable of summoning the appropriate place representation while estimating the agent's motion across space and time. The final row displays the ground truth trajectory, which aligns quite closely with the expectations of the cognitive map.

Finally, among all the reinforcement learning (RL) models employed in this study, an increase in the number of steps directly correlates with higher memory usage, frequently leading to failure if memory capacity falls short. In contrast, our approach provides a notably more efficient solution, requiring a maximum of 1G of memory space and mitigating scalability concerns related to the size of the environment. Refer to Table \ref{tab:testing_requirements} for a summary of the most demanding requirements for an exploration/goal task involving a maximum of 1500 steps.

\begin{table}[htb!]
\centering
\begin{tabular}{|c|c|c|c|}
\hline
model & n$^{\circ}$ CPU & n$^{\circ}$ GPU & \begin{tabular}[c]{@{}c@{}}used \\ Memory (G)\end{tabular} \\ \hline
Ours & 2 & 0 & 1 \\ \hline
Dreamerv3 & 2 & 1 & 28 \\ \hline
C-BET & 2 & 0 & 12 \\ \hline
RND & 2 & 0 & 9 \\ \hline
Curiosity & 2 & 0 & 11 \\ \hline
Count & 2 & 0 & 8 \\ \hline
\end{tabular}
\caption{Every model has specific system requirements, this table outlines the most demanding criteria needed to achieve successful exploration or goal seeking in the 4 by 5 rooms environment configuration.}
\label{tab:testing_requirements}
\end{table}

\end{document}